\documentclass{article}
\usepackage{ijcai19}

\usepackage{times}
\usepackage{soul}
\usepackage{url}
\usepackage[hidelinks]{hyperref}
\usepackage[utf8]{inputenc}
\usepackage[small]{caption}
\usepackage{graphicx}
\usepackage{amsmath}
\usepackage{booktabs}
\usepackage{algorithm}
\usepackage{algorithmic}
\usepackage{multirow}
\usepackage{subfigure}
\usepackage{amsfonts}
\usepackage{makecell}
\usepackage{xcolor}

\title{CGaP: Continuous Growth and Pruning for Efficient Deep Learning}

\author{
Xiaocong Du$^1$\footnote{Contact Author}\and
Zheng Li${^2}$\and
Yu Cao$^{1,2}$\\
\affiliations
$^1$Electrical Engineering, School of ECEE, Arizona State University\\
$^2$Computer Engineering, School of CIDSE, Arizona State University\\
\emails
\{xiaocong, zheng11, ycao\}@asu.edu
}

\begin{document}
\pagestyle{plain}

\maketitle

\begin{abstract}
Today a canonical approach to reduce the computation cost of Deep Neural Networks (DNNs) is to pre-define an over-parameterized model before training to guarantee the learning capacity, and then prune unimportant learning units (filters and neurons) during training to improve model compactness. We argue it is unnecessary to introduce redundancy at the beginning of the training but then reduce redundancy for the ultimate inference model. In this paper, we propose a Continuous Growth and Pruning (CGaP) scheme to minimize the redundancy from the beginning. CGaP starts the training from a small network seed, then expands the model continuously by reinforcing important learning units, and finally prunes the network to obtain a compact and accurate model. As the growth phase favors important learning units, CGaP provides a clear learning purpose to the pruning phase. Experimental results on representative datasets and DNN architectures demonstrate that CGaP outperforms previous pruning-only approaches that deal with pre-defined structures. For VGG-19 on CIFAR-100 and SVHN datasets, CGaP reduces the number of parameters by 78.9\% and 85.8\%, FLOPs by 53.2\% and 74.2\%, respectively;  For ResNet-110 On CIFAR-10, CGaP reduces 64.0\% number of parameters and 63.3\% FLOPs.

\end{abstract}

\section{Introduction}

\begin{figure}[t!]
\begin{center}
\includegraphics[width=\columnwidth]{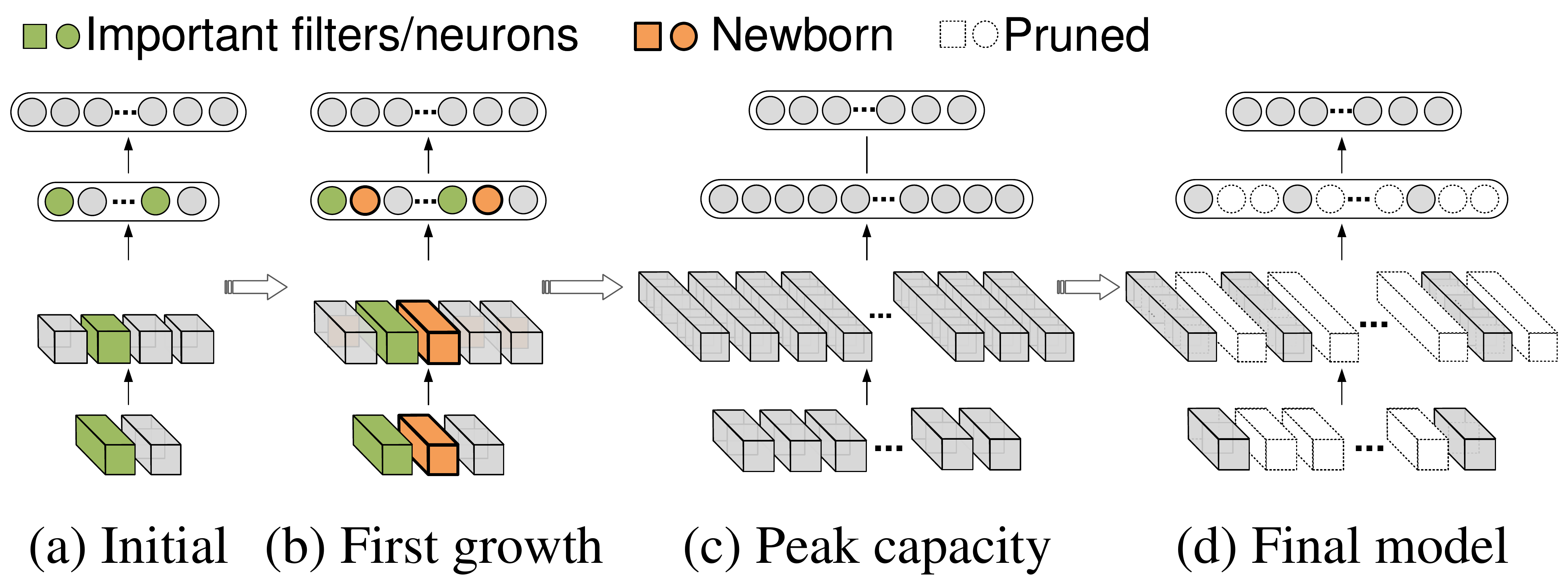}
\end{center}
\vspace{-0.2cm}
\caption{The flowchart of the proposed CGaP scheme. Different from a canonical pipeline, CGaP (a) initializes the learning from a small network seed instead of an over-parameterized structure, (b) continuously adds important learning units and expands each layer, (c) reaches peak capacity, (d) performs the pruning and eventually obtains a sparse structure with high accuracy.}
\label{fig:flow}
\end{figure}

Rapid development of DNNs has promoted various applications such as image classification~\cite{krizhevsky2012imagenet}, object detection~\cite{ren2015faster,zhang2016yin} and scene detection~\cite{zhou2017places,netzer2011reading}.  
However, the success of these tasks heavily relies on wider and deeper networks, making it increasingly difficult to deploy on resource-limited hardware platforms due to the excessive requirements of memory and computation cost. For example, 20.4 million parameters and 0.8 billion FLOPs are required by a typical VGG-Net~\cite{simonyan2014very} model to classify CIFAR-100 dataset~\cite{krizhevsky2009learning}. 
The intense computation and memory requirements of DNNs remain as a challenge for high-speed and low-power computing systems. 

One of the promising solutions to this challenge is the model compression. Recent works in this field can be categorized into three types:  low-rank decomposition, low-precision quantization, and network pruning. Among them, network pruning, a technique that removes parameters from an over-parameterized network, has been extensively researched to tackle network redundancy.
Network pruning is classified into \textit{saliency}-based pruning~\cite{han2015learning,haoli,hu2016network,liu2017learning,luo2017thinet}, and \textit{penalty}-based pruning~\cite{lebedev2016fast,wen2016learning}. \textit{Saliency}-based pruning removes unimportant weights or filters according to a predefined saliency score (a metrics to measure the impact on the training loss) and fine-tunes the rest.   \textit{Penalty}-based pruning penalizes weights by adding a regularization term into the loss function. These pruning schemes usually follow a three-step procedure: start training with a pre-defined large network, prune the trained model and fine-tune the pruned model to recover accuracy.

Nevertheless, the current pruning schemes have the following limitations: (1) Starting with an over-parameterized network could be sub-optimal as it introduces redundancy and overfitting, harming the inference accuracy and compactness. (2) During the training, current pruning schemes only discard secondary weights and filters but never strengthen important ones. This mechanism may hurt the learning efficiency. 

To tackle such limitations, we propose a novel scheme (Figure~\ref{fig:flow}), namely Continuous Growth and Pruning (CGaP).  CGaP starts training from a small seed, whose size is as low as 0.1\%-3\% of a full-size standard model.  Then CGaP continuously adds new filters and neurons according to our saliency score. Finally, a structured filter/neuron pruning is applied to the post-growth model, generating a significantly sparse structure while maintaining the accuracy. We argue that CGaP exceeds traditional three-step pruning scheme in efficiency owing to two advantages:  (1) Starting training from a small network reduces or even avoids the redundancy introduced at the very beginning and thus is less prone to overfitting. (2) The reinforcement of important learning units during the growth phase benefits the overall learning accuracy and the pruning phase as it provides a clearer training target.

The performance of CGaP is validated by the experimental results. For VGG-19, the proposed CGaP achieves $78.9\%$ parameter reduction with $+0.37\%$ accuracy improvement on CIFAR-100, $85.8\%$ parameter reduction with $+0.23\%$ accuracy improvement on SVHN~\cite{netzer2011reading}. For ResNet-110~\cite{he2016deep}, CGaP reduces $64.0\%$ parameters and $63.3\%$ FLOPs with $+0.09\%$ accuracy improvement on CIFAR-10. These results exceed the state-of-the-art pruning methods~\cite{han2015learning,hu2016network,haoli,liu2017learning,liu2018rethinking,he2018soft}. 

\paragraph{Contribution.} The contribution of this paper is three-fold: 
\textbf{(1)} A training flow (CGaP) with dynamic structure is proposed to alleviate over-parameterization without sacrificing the accuracy. The experiments are evaluated on standard image and scene datasets and the results exceed previous pruning-only pipelines in model size reduction, accuracy and training time.
\textbf{(2)} A novel saliency-based growth is introduced and the efficacy of our growth policy is validated by both mathematical proof and experiments. We further provide our understanding of the role that growth phase plays in the CGaP scheme.
\textbf{(3)} The proposed CGaP provides the ground for future design of adaptive networks for various dynamic tasks such as transfer, continual and lifelong learning.

\section{Related Work}
There has been intense interest in reducing the over-parameterization of DNNs.  The structure surgery has been widely used, including destructive and constructive methods. Destructive approaches remove connections or filters/neurons from the structure, generating sparse models. For instance, ~\cite{han2015learning} pruned connections by removing less important weights determined by magnitude. ~\cite{haoli} pruned individual filters layer by layer based on the saliency metrics of each filter.  Other similar saliency-based filter pruning approaches include~\cite{liu2017learning,he2017channel,hu2016network,guo2016dynamic}. Penalty-based sparsity regularization have been explored by~\cite{wen2016learning,liu2015sparse} and structured sparsity was achieved. Our method is different from all above pruning schemes from two perspectives:  We start training from a small seed other than an over-parameterized network. Apart from discarding unimportant filters/neurons, we also strengthen important ones to further raise accuracy and compactness. 

On the other hand, constructive methods add connections or neurons to increase capacity. \cite{ash1989dynamic,briedis1998using} enlarged network capacity with fresh neurons and evaluated rudimentary problems such as XOR problems and multi-layer perceptron (MLP), without datasets from a real scenario. Unlike their work, we expand the network with elaborately selected units and validate on advanced DNNs and datasets. ~\cite{dai2017nest} picked a set of convolutional filters from a bundle of randomly generated ones to enlarge the convolutional layers. However, how to find one set of filters that reduces the most loss among several sets is by trial-and-error method, consuming a significant amount of resources. Compared to it, directly growing up the convolutional layers from one seed, like CGaP does, is more efficient in real-world applications. To our knowledge, CGaP is the first scheme to employ continuous saliency-based growth for the purpose of efficient deep learning.

Orthogonal methods to our work include low-precision quantization and low-rank decomposition. Low-precision quantization~\cite{gong2014compressing,hubara2017quantized} quantized the parameter and gradients to fewer bits and thus reduced the memory size and access. \cite{denton2014exploiting,leng2018extremely} compressed each convolutional layer by finding an appropriate low-rank approximation. It is worth mentioning that our CGaP scheme can leverage the aforementioned orthogonal methods to further compress and accelerate the sparse model.

\section{Method}

\subsection{Terminology}
\label{section:terminology}
 A DNN can be treated as a feedforward multi-layer architecture that maps the input image to a certain output vector.  Each layer is a definite function, such as convolution, ReLU, pooling and inner product, whose input is $\mathcal{X}$, output is $\mathcal{Y}$ and parameter is $\mathcal{W}$ in case of convolutional and fully-connected layers.  The $l$-th convolutional layer (conv-layer) is formulated as: $\mathcal{Y}_l = \mathcal{X}_l\ast\mathcal{W}_l$, where $\mathcal{W}_l \in \mathbb{R}^{O_l\times I_l\times K\times K}$.  The $l$-th fully-connected layer (fc-layer) is represented by: $\mathcal{Y}_l = \mathcal{X}_l \cdot \mathcal{W}_l$, where $\mathcal{W}_l \in \mathbb{R}^{O_l \times I_l}$. 
    
 In conv-layers, the collection of weights that generates the $o$-th output feature map is denoted as  \textbf{filter} $W^o_l$,  $W^o_l\in\mathbb{R}^{I_l\times K\times K}$. The weight pixel at the coordinate of $(o,i,m,n)$ in conv-layers is denoted as $W_l^{o,i,m,n}\in\mathbb{R}^{1\times1}$.
 In fc-layers, the $i$-th hidden activation in the $l$-th layer is denoted as a \textbf{neuron}, $N^i_l$. This neuron receives input from the previous layer through its \textbf{fan-in} weights $W_{l,fan-in}^i\in\mathbb{R}^{1\times I_{l-1}}$ and propagates to the next layer through \textbf{fan-out} weights $W^i_{l,fan-out}\in\mathbb{R}^{O_l\times 1}$. $W_l^{o,i}\in\mathbb{R}^{1\times1}$ is a weight pixel at the coordinate of $(o,i)$ in fc-layers.

\paragraph{Learning units.} Growing or pruning a filter $W^o_l$ indicates adding or removing $W^o_l\in\mathbb{R}^{I_l\times K\times K}$ and its corresponding output feature map. Growing or pruning a neuron  $N^i_l$ means adding or removing both $W^i_{l,fan-out}$ and $W_{l,fan-in}^i$.

\begin{algorithm*}[t]
\caption{CGaP - Continuous Growth and Pruning scheme}
\label{CGaP-alg1}
\begin{algorithmic}[1] 
\STATE Initialize a small network model $M_{current}\leftarrow M_{initial}$.
\FOR{epoch = 1 to E}
    \STATE Train current model $M_{current}$ and fetch $Loss$ and $Accuracy$.
    \IF {$epoch\%\frac{1}{f_{growth}}=0$ and $M_{current}$ \textless $\tau_{capa.}$}
      \FOR{$l$ = 1 to L}
            \FOR{each filter $W^o_l$ in conv-layer $l$, or each neuron $N^i_l$ in fc-layer $l$}
             \STATE  Calculate growth score $GS_{W^o_l}$ according to Eq.~\ref{math:filter_score} and $GS_{N^i_l}$ according to Eq.~\ref{math:neuron_score}.
            \ENDFOR
            \STATE Sort all units and select $\beta O_l$ filters or $\beta I_l$ neurons with the highest $GS_{W^o_l}$ or $GS_{{N^i_l}}$.
            \FOR{j = 1 to $\beta O_l$ (for fc-layer, $\beta I_l$)}
                \STATE Add new units in layer $l$ and initialize with Eq.~\ref{math:initial1} and Eq.~\ref{math:initial2}.
                \STATE Map dimension in conv-layer $l+1$ (fc-layer $l-1$) and initialize with Eq.~\ref{math:initial3} and Eq.~\ref{math:initial4} (fc: Eq.~\ref{math:initial3-fc} and Eq.~\ref{math:initial4-fc}).
            \ENDFOR
      \ENDFOR
      \STATE $M_{current} \leftarrow M_{grown}$. 
      \ENDIF
    \IF {epoch$ \% \frac{1}{f_{pruning}}=0$ and $Accuracy>\tau_{accu.}$}
        \FOR{each weight $W_l^{o, i, m, n}\in\mathbb{R}^{1\times1}$ in conv-layer $l$ or each $W^{o,i}_l\in\mathbb{R}^{1\times1}$ in fc-layer $l$}
        \STATE Calculate weight pruning score $PS_W$ according to Eq.~\ref{math:PS1} for conv-layers and Eq.~\ref{math:PS2} for fc-layers.
        \ENDFOR
        \STATE Sort weights by $PS_W$.
        \STATE Zero-out the lowest $\gamma_W \prod(O_l, I_l, K, K)$ weights in conv-layer and $\gamma_W \prod(I_l, O_l)$ weights in fc-layer.
        \FOR{each filter $W^o_l$ (neuron $N^i_l$) in all layers}
           \STATE Zero-out entire filter $W^o_l$ (neuron $N^i_l$) if the weight sparsity is larger than pruning rate $\gamma_F$ ($\gamma_N$).
        \ENDFOR
        \STATE $M_{current} \leftarrow M_{pruned}$.
    \ENDIF
\ENDFOR
\STATE $M_{final} \leftarrow M_{current}$ and test $M_{final}$.
\end{algorithmic}
\end{algorithm*}

\subsection{Saliency Score} \label{section:score}
In this section, we provide mathematical formulation of the saliency score, which is used to measure the importance of each learning unit in the CGaP algorithm. The saliency score embodies the difference between the loss function with and without each unit. In other words, if the removal of a unit has a relatively small effect on the loss function, this unit is identified to be unimportant, and vice versa.

The objective function to get the filter with the highest saliency score is formulated as:
\begin{equation}
 \resizebox{0.99\linewidth}{!}{$
   \underset{W_l^o} {argmin} {|\Delta\mathcal{L}(W_l^o)|} \Leftrightarrow \underset{W_l^o} {argmin} |\mathcal{L}(\mathcal{Y};\mathcal{X},\mathcal{W}) \nonumber  \\ -\mathcal{L}(\mathcal{Y};\mathcal{X},W_l^o=\mathbf{0})|
$}
\end{equation} 

Using the first-order of the Taylor Expansion of $|\mathcal{L}(\mathcal{Y};\mathcal{X},\mathcal{W})-\mathcal{L}(\mathcal{Y};\mathcal{X},W_l^o=\mathbf{0})|$ at $W_l^o = \mathbf{0}$, we get:
\begin{align}\label{math:filter_score}   
   |\Delta\mathcal{L}(W_l^o)| &\simeq |\frac{\partial{\mathcal{L}}(\mathcal{Y};\mathcal{X},\mathcal{W})}{\partial{W_l^o}}W_l^o|  \nonumber\\ 
   & =\sum_{i=0}^{I_l}\sum_{m=0}^{K}\sum_{n=0}^{K}|\frac{\partial{\mathcal{L}}(\mathcal{Y};\mathcal{X},\mathcal{W})}{\partial{W_l^{o, i, m, n}}} W_l^{o, i, m, n}| 
\end{align} 

Similarly, the saliency score of a neuron is derived as:
\begin{eqnarray} \label{math:neuron_score}  
   |\Delta\mathcal{L}(N_l^i)| &\simeq& |\frac{\partial{\mathcal{L}}(\mathcal{Y};\mathcal{X},\mathcal{W})}{\partial{W_{l,fan-out}^i}}W_{l,fan-out}^i|  \nonumber\\  \textbf{}
   &=& \sum_{o=0}^{O_l}|\frac{\partial{\mathcal{L}}(\mathcal{Y};\mathcal{X},\mathcal{W})}{\partial{W_l^{o, i}}} W_l^{o, i}| 
\end{eqnarray}

\subsection{Continuous Growth and Pruning Flow} \label{section:Continuous Growth and Pruning flow}

\begin{figure}[t!]
\begin{center}
\includegraphics[width=\columnwidth]{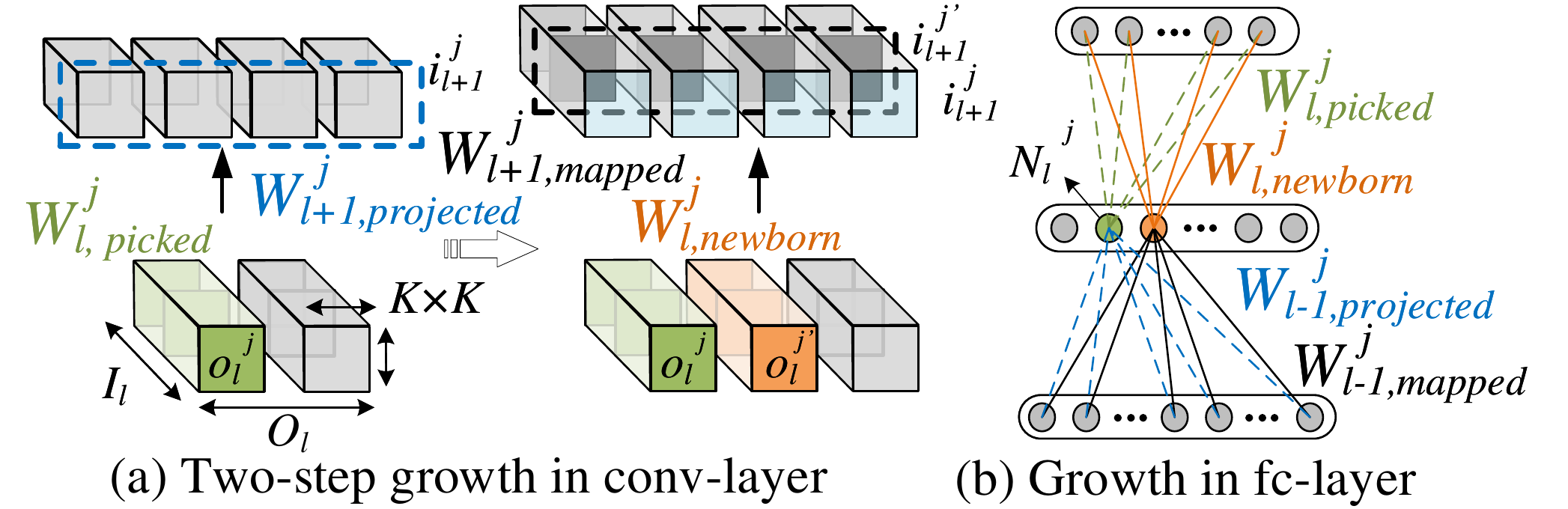}
\end{center}
\vspace{-0.2cm}
\caption{Illustration of two-step growth. In conv-layers, (a) the filter $W_{l, picked}^j$ (green) is selected according to the saliency score and  a new tensor $W_{l, newborn}^j$ (orange) is added. Then the input-wise tensor $W_{l+1, projected}^j$ (blue) in layer $l+1$ is projected, and $W_{l+1, mapped}^j$ (black) is generated. (b) In fc-layers, neuron's fan-out weights are added and fan-in weights are mapped.}
\label{fig:layer_wise}
\end{figure}

Based on the saliency score, we develop the entire CGaP flow, as presented in Algorithm \ref{CGaP-alg1}. The growth happens periodically at the frequency of $f_{growth}$ after network initialization and stops on reaching an intended capacity $\tau_{capa.}$. Using model capacity as the threshold helps guarantee the model size is within users' design requirement. The growth is exercised layer by layer from the bottom (input) to top (output) based on the local ranking of the saliency score. The pruning is then triggered if training accuracy reaches a threshold $\tau_{accu.}$, at which we consider the learning is approaching the end. Pruning is also executed locally in a layer-wise manner at the frequency of $f_{pruning}$.
Each iteration of growth consists of two steps: growth in layer $l$ and dimension mapping in the adjacent layer, as shown in Figure~\ref{fig:layer_wise}.

\paragraph{Growth in conv-layer $l$.} In the $t$-th growth, we first sort all the filters in this layer based on Eq.~\ref{math:filter_score}. Then we select $\beta O_{l,t}$ filters, where $\beta$ is the growth rate.  Adjacent to each selected filter $W_{l,picked}^j\in\mathbb{R}^{I_l\times K\times K}$ (Figure~\ref{fig:layer_wise}(a)), a tensor with the same dimension is created, namely $ W_{l,newborn}^j\in\mathbb{R}^{I_l\times K\times K}$. 
Ideally, we hope the new and old units cooperate with each other to optimize the learning performance. As the existing units have already learned on the data, to reconcile the learning pace of the new units and the pre-existing ones, we initialize the new units and scale the old ones following:
\begin{align}
\label{math:initial1}
W_{l,newborn}^j = \sigma W_{l,picked}^j + X \sim U([-\mu, \mu]) \\
\label{math:initial2}
W_{l,picked}^j = \sigma W_{l,picked}^j + X \sim U([-\mu, \mu])
\end{align}
where $\sigma$ $\in(0,1]$ is the scaling factor and $X$ is a random value following uniform distribution in $[-\mu, \mu]$, where $\mu$ $\in(0,1]$. Scaling is integral to growth as it reconciles the learning pace of old and new filters, prevents the exponential explosion of output due to the feedforward propagation $\mathcal{Y}_l = \mathcal{X}_l\ast\mathcal{W}_l$, and avoids over-fitting. The noise added prevents the model from a local minimum.

\paragraph{Mapping in conv-layer $l+1$.} As the number of filters in layer $l$ increases from $O_{l,t}$ to $(1+\beta)O_{l,t}$, the input-wise dimension in layer $l+1$ should also be increased from  $I_{l+1,t}$ to $(1+\beta)I_{l+1,t}$, where $I_{l+1,t}$= $O_{l,t}$, in order to maintain the consistency in dimension for the data propagation flow. We do so by locating the tensor $W_{l+1, projected}^j$, a projection of $W_{l, picked}^j$ in the adjacent layer, then create $W_{l+1,mapped}^j$ on the side. The value of $W_{l+1,mapped}^j$ and $W_{l+1,projected}^j$ are assigned following:
\begin{align}
\label{math:initial3}
W_{l+1,mapped}^j = \sigma W_{l+1,projected}^j + X \sim U([-\mu, \mu]) \\
\label{math:initial4}
W_{l+1,projected}^j = \sigma W_{l+1,projected}^j + X \sim U([-\mu, \mu])
\end{align}

After the two-step growth in layer $l$, layer $l+1$ grows and layer $l+2$ maps, so on and so forth till the last convolutional layer. It is worth to note that, for the skip-connections with $1\times1$ convolutions (`projection shortcuts'~\cite{he2016deep}) in ResNet~\cite{he2016deep}, the dimension mapping occurs between the two layers that the skip-connection links, not necessarily to be adjacent layers.

\paragraph{Growth and mapping in fc-layers.} See Figure~\ref{fig:layer_wise}(b). The neuron growth and scaling in fc-layers $l$ are similarly as conv-layers, following Eq.~\ref{math:initial1} and \ref{math:initial2}. The mapping in fc-layers take place in the fan-in weights following: 
\begin{align}
\label{math:initial3-fc}
W_{l-1,mapped}^j= \sigma W_{l-1,projected}^j + X \sim U([-\mu, \mu]) \\ 
\label{math:initial4-fc}
W_{l-1,projected}^j= \sigma  W_{l-1,projected}^j + X \sim U([-\mu, \mu])
\end{align}
When mapping the most bottom fc-layer with the most top conv-layer, we treat the flattened output feature map of the top conv-layer as the input from layer $l-1$ and map in the same way.

\paragraph{Pruning Phase.} Pruning in conv-layers and fc-layers are similar. We sort weight pixels in each conv-layer and fc-layer according to Eq.\ref{math:PS1} and Eq.\ref{math:PS2}, respectively: 
\begin{align}
\label{math:PS1}
PS_{W_l^{o, i, m, n}} = & |\frac{\partial{\mathcal{L}}(\mathcal{Y};\mathcal{X},\mathcal{W})}{\partial{W^{o, i, m, n}_l}}W_l^{o, i, m, n}|  \\
\label{math:PS2}
PS_{W_l^{o,i}} = &|\frac{\partial{\mathcal{L}}(\mathcal{Y};\mathcal{X},\mathcal{W})}{\partial{W_l^{o,i}}}W_l^{o,i}|
\end{align}
In each layer, 100$\gamma_W$\% (the weight pruning rate $\gamma_W$$\in (0,1)$) weight pixels with the lowest $PS_W$ are set as zero, followed by removing the entire filter (neuron) whose sparsity is larger than the filter (neuron) pruning rate $\gamma_F$ ($\gamma_N$)$\in(0,1)$.

\begin{table*}[t!]
\centering
\resizebox{0.86\textwidth}{!}{ 
\begin{tabular}[h]{c lrrrrrr}
\toprule[0.10em]
Model      & Method                             & Accuracy(\%)      & FLOPs                        & Pruned            & Parameter          & Pruned           & Speedup         \\ \midrule
\multirowcell{4}{LeNet-5\\on\\MNIST}    &  Baseline                          & 99.29             & $4.59\times10^6$             & --                & 431K            & --                &--        \\ \cmidrule[0.03em]{2-8}
                                        &  Pruning~\cite{hu2016network}      & -0.03             & $0.85\times10^6$             & 81.5\%            & 112K            & 74.0\%            &--              \\
                                        &  Pruning~\cite{han2015learning}    & -0.06             & $0.73\times10^6$             & 84.0\%            & 36K             & 92.0\%            &--              \\
                                        &  CGaP                              & \textbf{+0.07 }   & $\mathbf{0.44\times10^6}$    & \textbf{90.4\%}   & \textbf{8K}     & \textbf{98.1\%}   & $1.1\times$     \\ \midrule[0.08em]
\multirowcell{3}{VGG-16\\on\\CIFAR-10}  & Baseline                          &  93.25            & $6.30\times10^8$             & --                & 15.3M           & --                &--         \\\cmidrule[0.03em]{2-8}
                                        & Pruning~\cite{haoli}              & +0.15             & $4.10\times10^8$             & 34.9\%            & 5.4M            & 64.7\%            &--                \\
                                        & CGaP                              & \textbf{+0.34}    & $\mathbf{2.80\times10^8}$    & \textbf{56.2\%}   & \textbf{4.5M}   &\textbf{ 70.6\% }  &$1.3\times$        \\ \midrule
\multirowcell{4}{VGG-19\\on\\CIFAR-100}& Baseline                           &  72.63            & $7.97\times10^8$             & --                & 20.4M           & --                &--          \\\cmidrule[0.03em]{2-8}
                                       & Pruning~\cite{liu2018rethinking}   & -0.78             & NA                           & --                & 10.1M           & 50.5\%            &--                 \\
                                       & Pruning~\cite{liu2017learning}     & +0.22             & $5.01\times10^8$             & 37.1\%            & 5.0M            & 75.5\%            &--                 \\
                                       & CGaP                               & \textbf{+0.37}    & $\mathbf{3.73\times10^8}$    &  \textbf{53.2\%}  &\textbf{ 4.3M}   &\textbf{ 78.9\% }  & $1.4\times$        \\ \midrule
\multirowcell{3}{VGG-19\\on\\SVHN}     & Baseline                           & 96.02             & $7.97\times10^8$             & --                & 20.4M           & --                &--          \\\cmidrule[0.03em]{2-8}
                                       & Pruning~\cite{liu2017learning}     & +0.11             & $3.98\times10^8$             & 50.1\%            & 3.1M            & 84.8\%            &--                  \\
                                       & CGaP                               & \textbf{+0.23}    & $\mathbf{2.06\times10^8}$    & \textbf{74.2\%}   & \textbf{2.9M}   &\textbf{ 85.8\% }  & $1.6\times$    \\\midrule[0.08em]
\multirowcell{3}{ResNet-56\\on\\CIFAR-10}& Baseline                        &   93.03           & $2.68\times10^8$             &   --              & 0.85M           & --                & --          \\ \cmidrule[0.03em]{2-8}
                                      & Pruning~\cite{liu2018rethinking}   &   -0.47           & $1.82\times10^8$             & 32.1\%            & 0.73M           & 14.1\%            &--                 \\
                                      & CGaP                               & \textbf{+0.17}    & $\mathbf{1.81\times10^8}$    & \textbf{32.5\%}   & \textbf{0.53M}  & \textbf{37.6\%}   & $1.1\times$        \\\midrule
\multirowcell{4}{ResNet-110\\on\\CIFAR-10} & Baseline                      & 93.34             & $5.23\times10^8$             & --                 & 1.72M          & --                &--            \\\cmidrule[0.03em]{2-8}
                                      & Pruning~\cite{haoli}               & -0.23             & $3.10\times10^8$             & 40.7\%             & 1.16M          & 32.6\%            &--                    \\
                                      & Pruning~\cite{he2018soft}          & \textbf{+0.18}    & $3.00\times10^8$             & 40.8\%             & NA             & --                &--                    \\
                                      & CGaP                               & +0.09             & $\mathbf{1.92\times10^8}$    & \textbf{63.3\%}    & \textbf{0.62M} & \textbf{64.0\%}   & $1.1\times$          \\
\bottomrule[0.10em]
\end{tabular}
}
\vspace{-0.1cm}
\caption{Summary of the results. The ``Accuracy'' column denotes the testing accuracy for a baseline model (e.g., 99.29), the relative testing accuracy reported in the pruning-only papers (e.g., -0.03 means 0.03\% accuracy drop), and the relative testing accuracy achieved by the proposed CGaP approach (e.g., +0.07 means 0.07\% accuracy improvement). ``Speedup'' column denotes the acceleration achieved by CGaP in the training time as compared to [Han \textit{et al.},2015] under the same setting. `NA' means `not available' in the original paper.}
\label{table:overallresult}
\end{table*}

\section{Experiments}
\label{section:experiment}
We evaluate the proposed approach by performing experiments on advanced datasets and networks, and demonstrate: 
(1) CGaP largely reduces the model size (number of parameters), computation cost (number of FLOPs) and training time without sacrificing accuracy, exceeding previous pruning-only schemes.
(2) Saliency-based selective growth outperforms random growth in accuracy.
(3) The growth phase provides a clearer learning purpose to the pruning phase by reinforcing important units.

\subsection{Experimental Setup}

The experiments are performed with PyTorch~\cite{paszke2017automatic} on one NVIDIA GeForce GTX 1080 Ti platform. The code to reproduce the results will be made publicly available.

\paragraph{Datasets.} MNIST~\cite{lecun1998gradient} is a grey-scale 10-class handwritten digit dataset containing 60,000 training images and 10,000 test images. The CIFAR~\cite{krizhevsky2009learning} datasets consists of 60,000 (50,000 training images and 10,000 test images) $32\times32$ color images in 10 classes (CIFAR-10) and 100 classes (CIFAR-100). The Street View House Number (SVHN)~\cite{netzer2011reading} dataset is composed by $32\times32$ colored scene images with 73,257 training images and 26,032 test images. 

\paragraph{Network structures.} LeNet-5~\cite{lecun1998gradient} consists of two conv-layers and two fc-layers. VGG-16 and VGG-19~\cite{simonyan2014very} follow the same convolutional depth proposed in~\cite{simonyan2014very} but redesign with only two fc-layers to be fairly compared with pruning~\cite{haoli}. The structures of ResNet-56 and ResNet-110 are designed as same as~\cite{haoli}.  Each convolutional layer in VGG-Net and ResNet (except skip-connections) is followed by a batch normalization layer~\cite{ioffe2015batch}. During the training, the depth of the networks remains as a constant but CGaP varies the width of each layer.  While mentioning `baseline', we refer to the full-size model trained from scratch without sparsity regularization. While mentioning `pruning-only' works, we refer to the three-step pruning schemes that deal with static structures.

\paragraph{Hyper-parameters.} Standard Stochastic Gradient Descent with momentum $0.9$ and weight decay 5E-4 are used for training. The initial learning rate is set to 0.1, and is divided by 10 for every 30\% of the total training epochs. On MNIST, CIFAR-100 and SVHN datasets, we train 60, 220 and 100 epochs. On CIFAR-10, we train 200 epochs. 
Hyper-parameters in the growth phase are set as $\sigma=0.5, \mu=0.1, \tau_{capa.}=O_{1,baseline}$ (i.e., CGaP stops at the the $t$-th growth if $O_{1,t+1}>O_1^{baseline}$), $\beta=0.6, f_{growth}=3$ in the following experiments. Hyper-parameters in the pruning phase are set as $f_{pruning}=1, \tau_{accu.}=0.9$.

\subsection{Performance Evaluation}

\paragraph{Overall Results.} In Table~\ref{table:overallresult}, we summarize the accuracy, number of FLOPs, and number of parameters of the efficient model attained by CGaP.  The calculation of FLOPs follows~\cite{molchanov2016pruning}. CGaP achieves superior performance than  previous pruning-only schemes that deal with pre-defined structures. 
For example, compared to the baseline VGG-19 on CIFAR-100,~\cite{liu2017learning} reduced the number of FLOPs to $5.01\times10^8$ with +0.22\% accuracy. CGaP outperforms it by reducing the number of FLOPs to $3.73\times 10^8$, with +0.37\% accuracy improvement. On excessively deep networks like ResNet-110, CGaP further reduce 63.3\% FLOPs compared to 40.8\% reported by~\cite{he2018soft}. The significant reduction in parameters and FLOPs is because CGaP prunes considerable amounts of entire filters.
Moreover, CGaP speeds up the training process. For example, CGaP achieves $1.6\times$ speedup in the run-time until convergence on SVHN. The acceleration is by virtue of the lower fine-tuning efforts required by CGaP compared to pruning-only schemes. In the growth phase, some units are labeled as important ones; these units are less prone to be pruned in the pruning phase, lessening the fine-tuning.

\begin{figure}[t!]
\begin{center}
\includegraphics[width=\columnwidth]{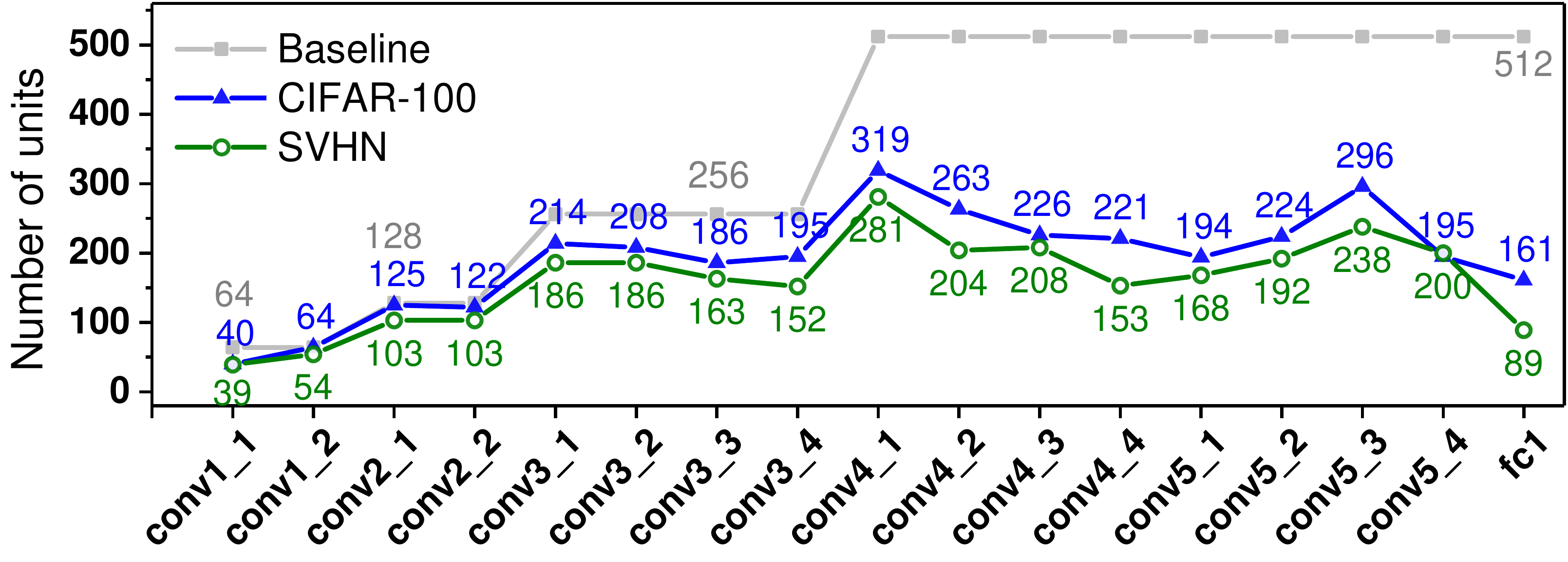}
\end{center}
\vspace{-0.2cm}
\caption{The final VGG-19 structure learned by CGaP. The top layers (from layer `conv4\_2' to layer `fc1') are \textless60\% in the size as compared to the baseline model.}
\label{fig:layer_size}
\end{figure}

\begin{figure}[t!]
\begin{center}

\includegraphics[width=\columnwidth]{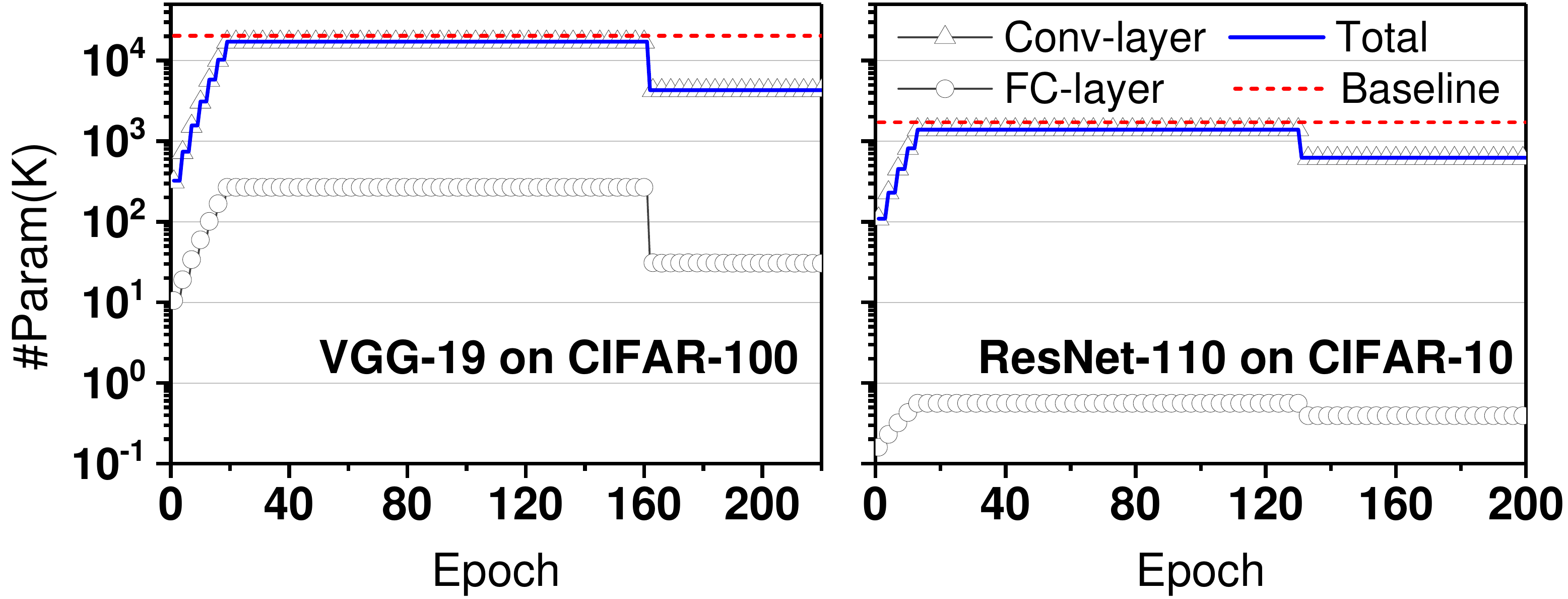}
\end{center}
\vspace{-0.2cm}
\caption{Number of parameters during training. During the growth phase, the model size continuously increases and reaches a peak capacity. When the pruning phase starts, the model size drops.}
\label{fig:paramVsEpoch}
\end{figure}
\begin{figure}[t!]
\begin{center}
\includegraphics[width=\columnwidth]{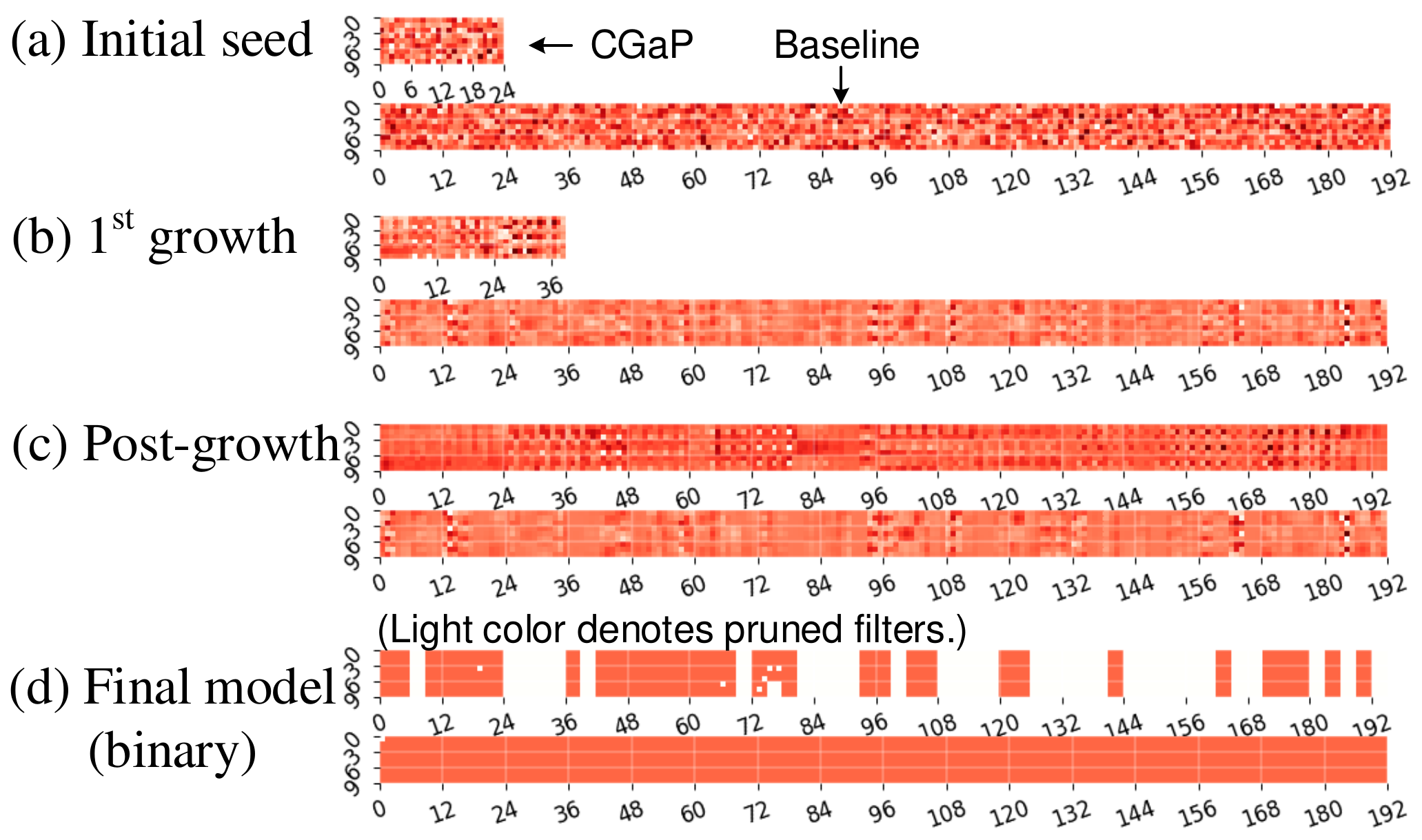}
\end{center}
\vspace{-0.2cm}
\caption{Visualization of the weights in conv1\_1 in VGG-19 on CIFAR-100 at four specific moments (a-d). Top bar: CGaP; bottom bar: baseline.  X-axis is the index of output-wise weights and Y-axis is the index of input-wise weights. The number of filters from (a) to (d) in the CGaP model are $8\rightarrow13\rightarrow65\rightarrow33$, while it is always 64 in the baseline.}
\label{fig:heatmap}
\end{figure}

\paragraph{Learning architecture.} CGaP not only decreases the model size, but also gains structured models. For LeNet-5, CGaP learns a final structure of [8-17-23-10] (number of filters/neurons in [conv1-conv2-fc1-fc2]), which is much smaller than the baseline model [20-50-500-10]. Another example is the VGG-19 structure learned by CGaP, as exhibited in Figure~\ref{fig:layer_size}.  In  the  baseline  model, the top layers  are  usually  pre-designed to  have  more  filters  than the bottom  layers. However, CGaP implies that it is not always necessary for the top layers to be wider. Figure~\ref{fig:paramVsEpoch} provides two examples of how the model size changes during the CGaP process. CGaP has a dynamic capacity in the training,  and the final model is significantly smaller than the baseline.  These results reveal the redundancy of a pre-designed network and prove CGaP is a compelling approach to reduce redundancy.

\paragraph{Understanding the growth.}  Figure~\ref{fig:heatmap} exhibits a visualization of weights to help understand the role that the growth phase plays in the CGaP. At initialization (a), the weights present a uniform distribution in both CGaP and the baseline since they are initialized randomly. At (b), the CGaP model begins the growth, from when the weight distribution starts to unveil a clearer structured pattern than the baseline, meaning the filters are fetching feature maps more effectively. After several iterations of growth, the model at (c) reveals a well-structured pattern. After pruning (d), the final model is in structured sparsity, which is more hardware-efficient than non-structured sparsity as discussed in~\cite{haoli,wen2016learning}.
From (c) to (d), we observe that most of the growth-favored filters (e.g., at x= 36, 48, 72, 96 in (c)) are retained after pruning. In other words, even after a long training between the post-growth and pre-pruning, most of the growth-labeled filters are still labeled as important ones. These observations imply that the growth phase benefits the overall learning accuracy and the pruning phase by providing a clearer learning target.

\begin{figure}[t!]
\begin{center}
\includegraphics[width=\columnwidth]{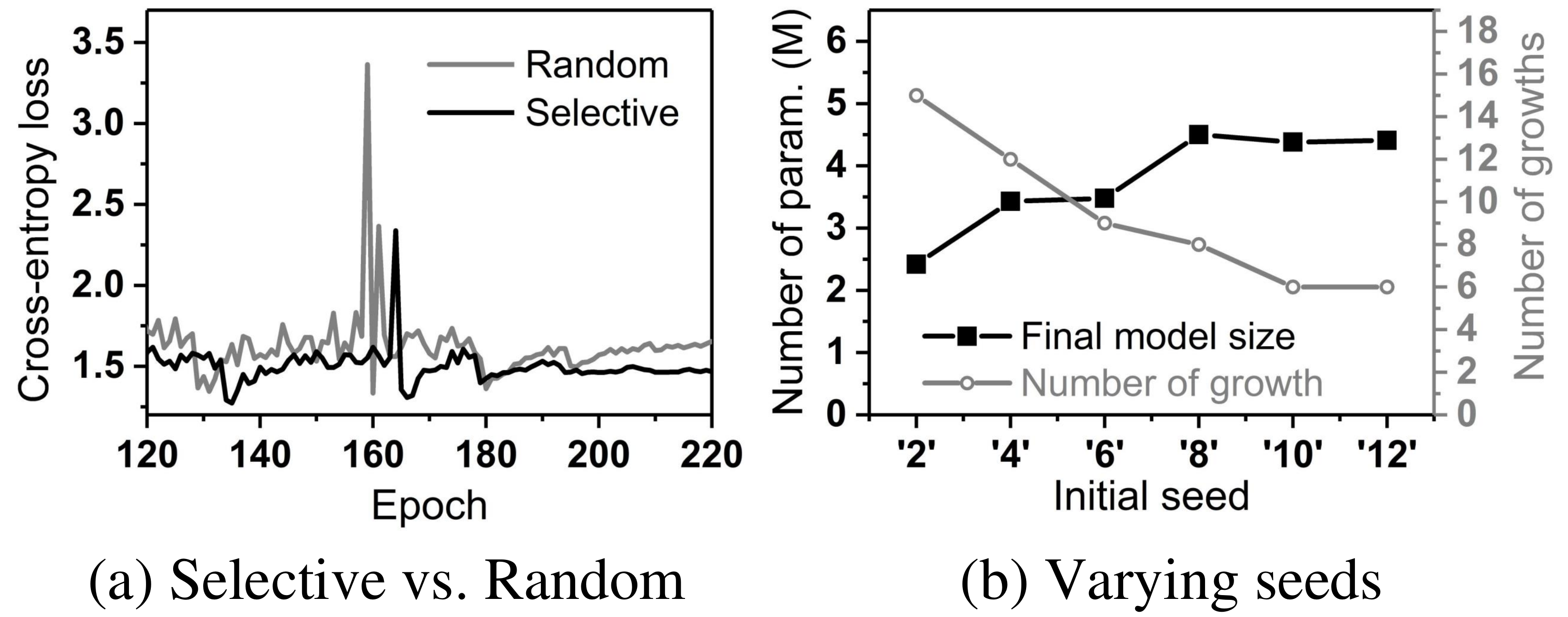}
\end{center}
\vspace{-0.2cm}
\caption{(a) Saliency-based growth outperforms random growth. (b) A larger seed leads to a larger final model but fewer iterations in the growth phase. Both figures are from CIFAR-100 results.}
\label{fig:graph2}
\end{figure}

\paragraph{Selective vs. random growth.} Figure~\ref{fig:graph2}(a) illustrates the efficacy of the saliency score as discussed in Section~\ref{section:score} from two aspects: (1) Selective growth, which emphasizes important units, has lower cross-entropy loss than random adding some units. (2) The spiking of loss caused by the first iteration of pruning in selective growth is  $1.4\times$ lower than that in random growth. These support the aforementioned argument that the growth phase benefits the following pruning phase.

\subsection{Ablation Study}\label{section:understanding}

\paragraph{Robustness of the seed.} As presented in Figure~\ref{fig:graph2}(b), when  the initial seed size increases from 0.01M (seed `2', $O_{1,initial}$=2) to 0.53M (seed `12',  $O_{1,initial}$=12), the final model also increases but requires fewer iterations of growth. Meanwhile, the accuracy is robust under the variation of seeds: relative test accuracy is -0.69\%, -0.2\%, -0.16\%, +0.37\%, +0.04\%, 0.29\% for seed `2'`to `12'.

\paragraph{Robustness of the hyper-parameters.}
CGaP hinges on a set of growth parameters to achieve an optimal architecture but the accuracy is stable under the change of these parameters. Variation in accuracy is \textless$2\%$ for an unoptimized set of hyper-parameters. Heuristically, we use the following intuitions to perform parameter optimization: a larger growth rate $\beta$ for a smaller seed and vice versa; threshold $\tau_{capa}$ can be set based on the user's intended model size; a larger $f_{growth}$ for a simpler dataset and vice versa; a greedy growth (large $\beta$ and $f_{growth}$) prefers a large noise $\mu$ but small $\sigma$ to push the model away from local minimum.  Tuning of the pruning rate is performed in a manner similar to~\cite{haoli}.

\section{Conclusion and Future Work}
In this paper, we propose a dynamic CGaP algorithm to reduce the computation cost of modern DNNs without accuracy loss. CGaP initializes the training from a small network, expands the network continuously with important learning units, followed by pruning unimportant ones. Experimental results demonstrate that CGaP achieves competitive performance compared to the state-of-the-art pruning-only schemes. Our approach and analysis will further inspire and support the development of adaptive neural networks for dynamic tasks such as continual and lifelong learning.

\section*{Acknowledgments}
This work was supported in part by C-BRIC, one of six centers in JUMP, a Semiconductor Research Corporation (SRC) program sponsored by DARPA. It was also partially supported by National Science Foundation (NSF) CCF \#1715443.

\bibliographystyle{named}
\bibliography{ijcai19}

\end{document}